\documentclass[a4paper,10pt,conference]{ieeeconf} % Comment this line out if you need a4paper
\usepackage{cite}
\usepackage{amsmath,amssymb,amsfonts}
\usepackage{algorithmic}
\usepackage{graphicx}
\usepackage{textcomp}
\usepackage{xcolor}
\usepackage{float}
\usepackage{url}
\usepackage{booktabs}
\usepackage{caption}
\usepackage{stfloats}% lets star‑floats appear at the top of the page

\IEEEoverridecommandlockouts                              % This command is only needed if 
\usepackage{url}
\usepackage{subfig}
\usepackage{graphicx}
\usepackage{varioref}
\graphicspath{{figures/}}
\labelformat{equation}{\textup{(#1)}}
\usepackage[pageanchor=true,
    plainpages=false,
    pdfpagelabels,
    bookmarks,
    bookmarksnumbered,
    pdftex,	
    pdfstartview=FitH,
    linkcolor=black,
    citecolor=black,
    urlcolor=black,
    filecolor=black]{hyperref}
\usepackage{amssymb}
\usepackage{multirow}
\usepackage{todonotes}
\usepackage{flushend}

\usepackage{array}
\newcolumntype{L}[1]{>{\raggedright\let\newline\\\arraybackslash\hspace{0pt}}m{#1}}
\newcolumntype{C}[1]{>{\centering\let\newline\\\arraybackslash\hspace{0pt}}m{#1}}
\newcolumntype{R}[1]{>{\raggedleft\let\newline\\\arraybackslash\hspace{0pt}}m{#1}}

\def\usenatbib{1}
\ifx\usenatbib\undefined
    \usepackage{cite}
\else
    \makeatletter
    \let\NAT@parse\undefined
    \makeatother
    \usepackage[numbers,sort]{natbib}

    \setcitestyle{citesep={], [}}
    \makeatletter
    \def\NAT@def@citea{\def\@citea{\NAT@separator}}%
    \makeatother
\fi

\graphicspath{ {./images/} }

\let\orgautoref\autoref
\providecommand{\Autoref}
        {\def\equationautorefname{Equation}%
         \def\figureautorefname{Figure}%
         \def\subfigureautorefname{Figure}%
         \def\Itemautorefname{Item}%
         \def\tableautorefname{Table}%
         \def\exerciseautorefname{Exercise}%
         \def\starexerciseautorefname{Exercise}%
         \def\sectionautorefname{Section}%
         \def\subsectionautorefname{Section}%
         \def\subsubsectionautorefname{Section}%
         \def\chapterautorefname{Section}%
         \def\partautorefname{Part}%
         \orgautoref}

\renewcommand{\autoref}
        {\def\equationautorefname{Equation}%
         \def\figureautorefname{Fig.}%
         \def\subfigureautorefname{Fig.}%
         \def\Itemautorefname{item}%
         \def\tableautorefname{Table}%
         \def\exerciseautorefname{Exercise}%
         \def\starexerciseautorefname{Exercise}%
         \def\sectionautorefname{Section}%
         \def\subsectionautorefname{Section}%
         \def\subsubsectionautorefname{Section}%
         \def\chapterautorefname{Section}%
         \def\partautorefname{Part}%
         \orgautoref}

\usepackage{hyperref}
\hypersetup{
colorlinks=true,
urlcolor=red,
}
\title{\LARGE \bf
CA-Cut: Crop-Aligned Cutout for Data Augmentation to Learn More Robust Under-Canopy Navigation
}

\author{Robel Mamo$^{1}$ and Taeyeong Choi$^{1}$% <-this % stops a space
\thanks{$^{1}$Robel Mamo and Taeyeong Choi are with the Learning and SEnsing Research (LaSER) laboratory at Kennesaw State University, Marietta, GA $30060$, USA. 
        {\tt\small rmamo@students.kennesaw.edu}, {\tt\small tchoi3@kennesaw.edu}}%
}

\begin{document}

%%%%%%%%%%%%%%%%%%%%%%%%%%%%%%%%%%%%%%%%%%%%%%%%%%
% for arXiv
\onecolumn
\vspace*{\fill}
    © 2025 IEEE. Personal use of this material is permitted. Permission from IEEE must be obtained for all other uses, in any current or future media, including reprinting/republishing this material for advertising or promotional purposes, creating new collective works, for resale or redistribution to servers or lists, or reuse of any copyrighted component of this work in other works.
\vspace*{\fill}
\twocolumn
\clearpage
%%%%%%%%%%%%%%%%%%%%%%%%%%%%%%%%%%%%%%%%%%%%%%%%%%

\maketitle
\thispagestyle{empty}
\pagestyle{empty}

%%%%%%%%%%%%%%%%%%%%%%%%%%%%%%%%%%%%%%%%%%%%%%%%%%%%%%%%%%%%%%%%%%%%%%%%%%%%%%%%
\begin{abstract}

State-of-the-art visual under-canopy navigation methods are designed with deep learning-based perception models to distinguish traversable space from crop rows. 
While these models have demonstrated successful performance,  
they require large amounts of training data to ensure reliability in real-world field deployment.
% from the fields that they are going to be deployed in. 
% Though there has been publicly available datasets, 
% they are very limited in number and in capturing the variety of challenges that are encountered by models deployed for real-time task execution. 
However, data collection is costly, demanding significant human resources for in-field sampling and annotation. 
% Although current methods use augmentation techniques such as 
To address this challenge, various data augmentation techniques are commonly employed during model training, such as 
color jittering, Gaussian blur, and horizontal flip, to diversify training data and enhance model robustness. 
% gain significant improvements in model robustness, 
% they fail to address challenges such as row spacing variability, crop growth levels, and occlusion. 
In this paper, we hypothesize that utilizing only these augmentation techniques may lead to suboptimal performance, particularly in complex under-canopy environments with frequent occlusions, debris, and non-uniform spacing of crops. 
Instead, we propose a novel augmentation method, so-called Crop-Aligned Cutout~(\mbox{CA-Cut})---which \emph{masks} random regions out in input images that are spatially distributed ``around'' crop rows on the sides---to encourage trained models to capture high-level contextual features even when fine-grained information is obstructed. 
% learn more robust perception models for under-canopy navigation. 
% particularly for under-canopy navigation with occlusions and varying gaps between crops. 
Our extensive experiments with a public cornfield dataset demonstrate that masking-based augmentations are effective for \emph{simulating} occlusions and significantly improving robustness in semantic keypoint predictions for visual navigation. 
In particular, we show that biasing the mask distribution toward crop rows in \mbox{CA-Cut} is critical for enhancing both prediction accuracy and generalizability across diverse environments---achieving up to a $36.9\%$~reduction in prediction error.  
% the performance of \mbox{CA-Cut} can be 
% maximized through \emph{curriculum} training---i.e.,~dynamically controlling the proximity of generated masks to the crops over training epochs. 
In addition, we conduct ablation studies to determine the number of masks, the size of each mask, and the spatial distribution of masks to maximize overall performance. 
The source code is publicly available at \url{https://github.com/mamorobel/CA-Cut}.
% We find that introducing a combination of naive uniformly distributed cutouts and label informed cutouts with Gaussian noise further improves model robustness by mitigating the challenges not resolved by current practices. 
% We take an evidence-based iterative approach to show the effects of crop-aligned cutouts on model performance.

\end{abstract}

%%%%%%%%%%%%%%%%%%%%%%%%%%%%%%%%%%%%%%%%%%%%%%%%%%%%%%%%%%%%%%%%%%%%%%%%%%%%%%%%
\section{Introduction}
\label{sec:intro}

% There is a growing need of robotics in various fields as we move towards a future of automation like we have never seen before. A rapidly growing field, in the context of automation and robotics, is agriculture. With this growth in need for automation in agriculture, come complexities that do not arise in other more structured environments, including but not limited to: inconsistencies in crop spacing, crop growth level, occlusion, row spacing, limited labeled and structured data availability, and many more. These problems "force" us to design solutions that generalize over the scope of the dataset that we have available. In this paper, we survey the impacts of applying naive random erasing during training time and test to see if we can make informed erasing to create a more robust and generalizable model. By comparing the baseline model against random erase augmentation during training, we see that there is a positive impact on model performance. This work is done on the visual navigation model developed by cropfollow++ [cite]. 
% ------
% some facts:
% the field of agricultural robotics has been rapidly growing rapidly
% their is a lack of datasets that are publicly available to train robust models that are able to handle the challenging agricultural settings
% the lack of available public datasets creates a need to explore different augmentation techniques
% common augmentations include horizontal flip, color jittering, and gaussian blur.
% However, this augmentations are not solving complex problems such as row spacing and occlusion. 

\begin{figure}[t]\centering
    \subfloat[]{\label{fig:ex_occlusion}%
    \includegraphics[width=.47\linewidth]{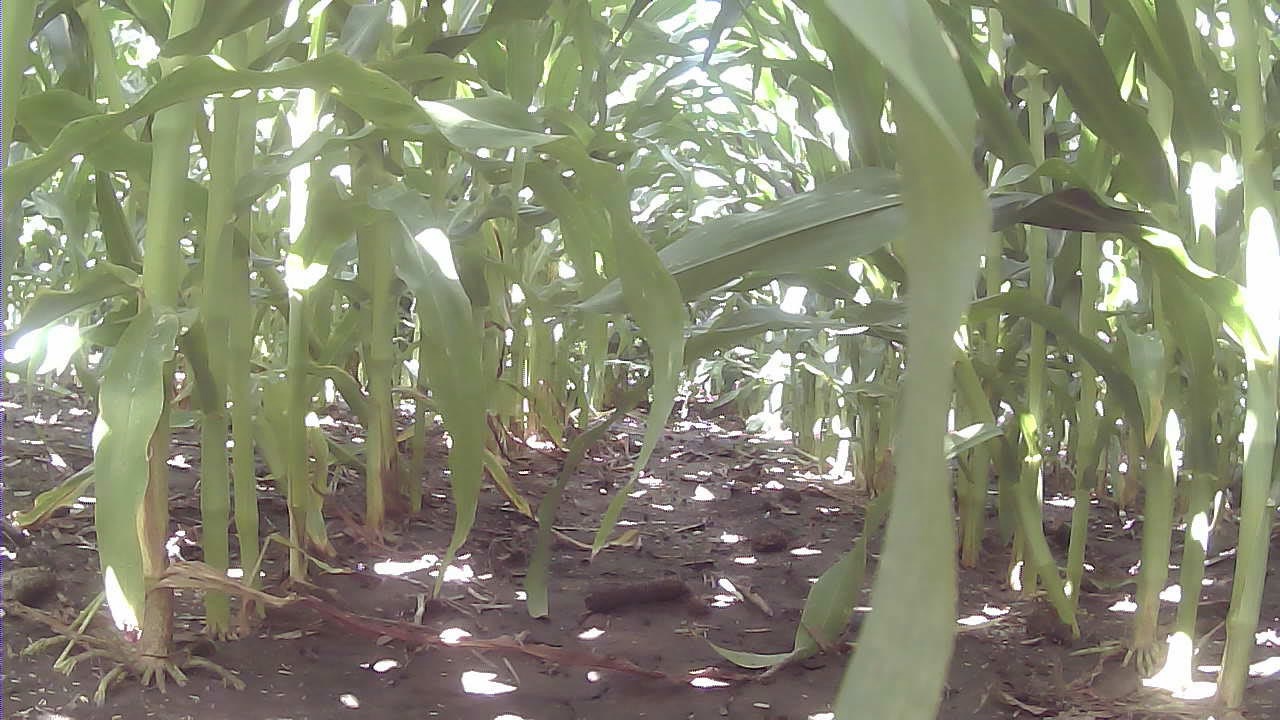}}
    \quad
    \subfloat[]{\label{fig:ex_spacing}%
    \includegraphics[width=.47\linewidth]{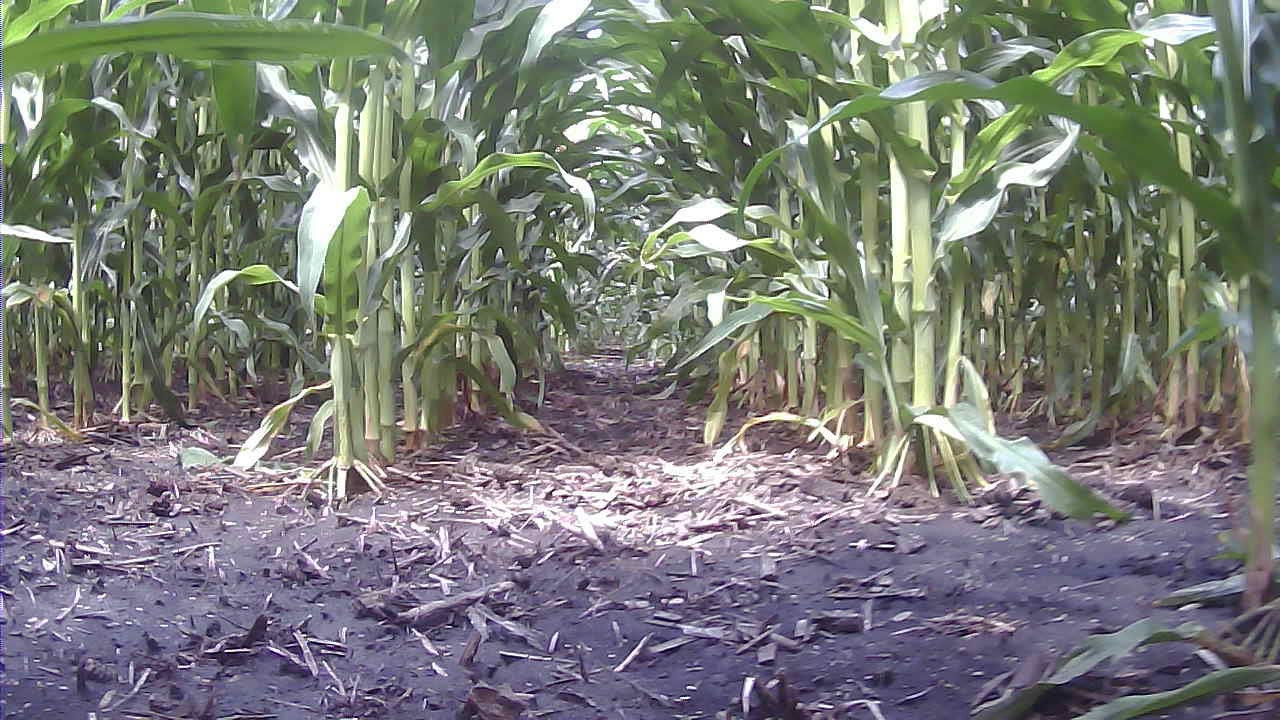}}
        \caption{
            Examples from the CropFollow dataset~\citep{SMGEVCG21}: \protect\subref{fig:ex_occlusion} severe occlusion, \protect\subref{fig:ex_spacing} debris and wide spacing in crop rows. 
            }
        \label{fig:extreme_examples}
\end{figure}

Reliable perception capability is essential for Unmanned Ground Vehicles~(UGVs) to safely navigate through agricultural fields and perform tasks for precision agriculture. 
Visual navigation is, however, particularly challenging in under-canopy environments due to factors such as frequent occlusions, debris, and variable spacing of crops (cf.~\autoref{fig:extreme_examples}). 
While deep learning-based methods have been widely adopted to enable perception models to effectively distinguish traversable space from surrounding crops~\citep{sivakumar2024demonstrating, WCZLZ22}, they require large amounts of training data to ensure reliability in real-world field deployment. 
Unfortunately, data collection is costly, demanding considerable human resources for in-field sampling and annotation. 

\begin{figure}[t]\centering
    \subfloat[Cutout]{\label{fig:concept_cutout}%
    \includegraphics[width=.47\linewidth]{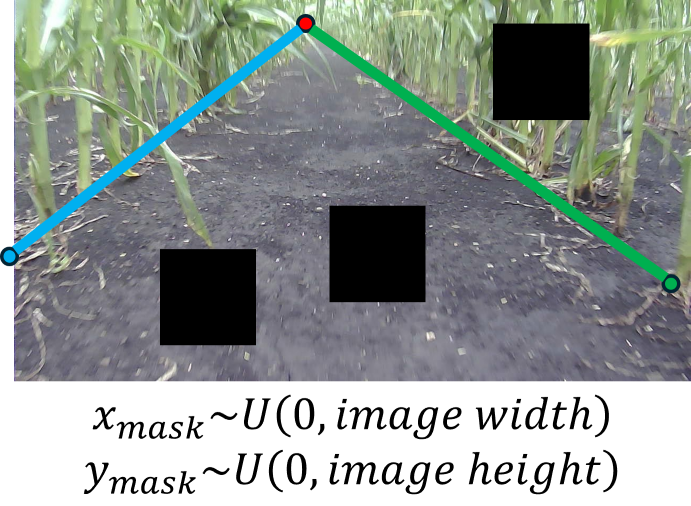}}
    \quad
    \subfloat[CA-Cut (Ours)]{\label{fig:concept_cacut}%
    \includegraphics[width=.47\linewidth]{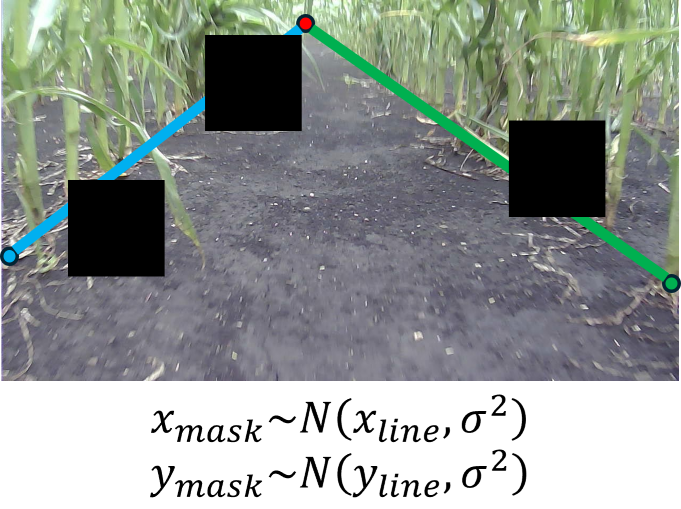}}
        \caption{
        Conceptual illustration of augmented images by Cutout~\citep{DT17} and CA-Cut in the semantic keypoint prediction task~\citep{sivakumar2024demonstrating}, where $x_{mask}$ and $y_{mask}$ represent the randomly sampled central $x$- and $y$-coordinates of each mask, and $x_{line}$ and $y_{line}$ denote random $x$- and $y$-coordinates sampled on either the blue or green line.  
        Three circles correspond to the vanishing point~(red), left~(blue), and right~(green) intercepts. 
            }
        \label{fig:concept}
\end{figure}

To address this issue, common image augmentation techniques (e.g.,~color jittering, Gaussian blur, and horizontal flip) are typically applied during model training to ensure robust performance on diversified data~\citep{sivakumar2024demonstrating, DCG22}. 
However, in this paper, we hypothesize that relying solely on these techniques may result in suboptimal perception models for under-canopy navigation. 
Furthermore, we argue that additional manipulations to \emph{simulate} occlusions and structural variability need to be considered in augmenting images. 
Inspired by Cutout~\citep{DT17}, we introduce Crop-Aligned Cutout~(\mbox{CA-Cut}), designed to \emph{mask} random regions in input images while distributing them ``around'' crop rows on the sides (\autoref{fig:concept_cacut}). 
In particular, this approach aims to encourage trained models to capture ``high-level'' contextual features for spatial understanding of crop rows, even in the absence of fine-grained cues of crops. 
To the best of our knowledge, we are the first to develop masking-based data augmentation techniques for agricultural visual navigation, although it has been often employed in conventional computer vision tasks~\citep{CK24,LSYP21}, including agricultural object detection~\citep{ZLMWW22,SOB25}.  

As shown in~\autoref{fig:concept}, we focus on the \emph{semantic keypoint prediction} problem~\citep{sivakumar2024demonstrating} in under-canopy cornfield environments, detecting three keypoints---the vanishing point of the crop row lines, the intercept of the left crop line with the boundary of the image, and the counterpart of the right crop line. 
In~\citep{sivakumar2024demonstrating}, the authors demonstrated the utility of the predicted keypoints to infer critical information for safe navigation, including the robot's heading and the distance ratio between crops. 
To be specific, our experiments with the CropFollow dataset~\citep{SMGEVCG21} demonstrate that masking-based augmentations (e.g.,~Cutout~\citep{DT17} and \mbox{CA-Cut}) effectively simulate occlusions, enabling perception models to learn robust features for keypoint prediction in visual navigation and outperform ones using only traditional augmentations.
In addition, we show that \mbox{CA-Cut}'s masks, \emph{aligned} with crop row lines, play a crucial role in improving both prediction accuracy and generalizability across diverse environments. 
% the performance of CA-Cut can be maximized through curriculum training---i.e.,~dynamically changing the proximity of generated masks from the crops over training epochs. 
Moreover, we perform ablation studies to explore different hyperparameters (i.e.,~the number of masks, the size of each mask, and the spatial distribution of masks) to maximize overall performance. 

\section{Related Works}

\subsection{Visual Navigation in Agricultural Fields}
\label{sec:visual_navigaton}

% Visual navigation under crop canopies presents unique challenges due to occlusions, irregular crop spacing, and clutter—all of which interfere with the perception systems of ground robots. While deep learning-based models have emerged as powerful tools for this task, their effectiveness hinges on access to large, annotated datasets, which are time-consuming and labor-intensive to collect in agricultural settings.

Visual navigation has been widely explored in the agricultural robotics community to enable autonomous crop monitoring in various crops such as corn~\citep{sivakumar2024demonstrating, SMGEVCG21}, sugar beet~\citep{DCG22, DCWG24}, and vineyards~\citep{ACMC21}. 
Recent approaches typically train U-Net-based neural networks~\citep{ronneberger2015u} on RGB~images captured by forward-facing monocular cameras to segment crop rows and generate a traversable reference path for UGVs---either over the canopy~\citep{DCG22, DCWG24} or between rows~\citep{ACMC21}, depending on crop height or task requirements.
These vision-based systems eliminate the need for additional sensors such as LiDAR, enabling low-cost, energy-efficient deployments. 

Instead of segmentation-based methods, the authors in~\citep{sivakumar2024demonstrating} proposed a new type of task---localizing three semantic keypoints (i.e.,~vanishing, left/right intercept points) in each image---particularly tailored for under-canopy navigation, where color-based pixel-wise segmentation often fails due to occlusions and visual clutter. 
Their framework, called \mbox{CropFollow++}, used a control module that inferred the robot's pose based on the predicted semantic keypoints and generated optimal linear/angular velocities to track reference paths.
Its effectiveness was demonstrated using physical mobile robots in corn fields with minimal human intervention. 

In this paper, we also investigate robot visual navigation in cornfields, leveraging the public dataset from~\citep{sivakumar2024demonstrating, SMGEVCG21}. 
However, we introduce a novel image \emph{augmentation} method, \mbox{CA-Cut}, aimed at improving the perception module by enhancing robustness in semantic keypoint predictions---ultimately benefiting downstream control modules.

% Recent efforts have aimed to reduce reliance on extensive field data by leveraging structurally meaningful representations and simulation-based training. CropFollow++ introduced a keypoint-based perception system for under-canopy navigation, which uses the vanishing point and crop row intercepts to infer robot orientation and lateral offset. This modular and interpretable approach enabled robust navigation across varied field conditions, including heavy occlusions and uneven terrain, and reduced failures compared to previous end-to-end learning systems \cite{sivakumar2024demonstrating}.

\subsection{Image Augmentation with Masks}
\label{sec:augmentation_techniques}

% many refs 
Random mask-based image augmentations have shown successful results with improved model generalization  in computer vision tasks. 
Cutout~\citep{DT17}, for instance, was to augment images by applying fixed-size zero masks at random locations in an image and demonstrated consistent improvement across image recognition models. 
Similarly, CutPaste~\citep{LSYP21} (CutMix~\citep{YHOCCY19}) augements input images by embedding rectangular masks, where instead of zeroing out regions, patches from a random location within the same image (another training image) are pasted into different regions. 
The authors used this technique to learn structure-sensitive representations for defect detection on industrial items in a self-supervised manner, leveraging a pretext task to classify images with CutPaste from the original. 
More recently, Colorful Cutout~\citep{CK24} has been introduced, using multi-colored segments within each mask and gradually increasing mask complexity with more colored segments across training epochs to perform curriculum learning. 

Inspired by all these, \mbox{CA-Cut} is also designed to erase the content within randomly selected regions to simulate partial occlusion of visual features in training images. 
% Thus, during training, the neural network is forced to learn using contextual information instead of focusing on fine-grained features.  
However, the key distinction in our work is that the placements of masks are biased toward a certain type of content (i.e.,~crop rows) which is considered to be visually informative. 
Moreover, our target application domain is mobile robot navigation, where learning robust perception is particularly critical.   

\subsubsection{Agricultural Scenarios}

% While occlusion is one of the biggest challenges in agricultural vision tasks, 
Compared to traditional augmentation methods (e.g., color jittering, flipping, and blur)~\cite{CWSC22, sivakumar2024demonstrating}, 
masking-based augmentation remains underexplored in agricultural vision.
For example, the authors in~\citep{BVVK21} enhanced broccoli head detectors to be robust to occlusion by acquiring training images with intentional partial occlusions by \emph{natural} leaves during data collection. 
While this approach yielded useful samples, the manual setup limited the scale of augmentation. 
In~\citep{RVK22}, cow pose estimation models were developed and evaluated on video frames with zero-value masks that removed parts of cow pixels to test robustness against partial occlusion. 
Yet, the training itself did not involve such masked images. 
In~\citep{ZLMWW22}, Cutout~\citep{DT17} and CutMix~\citep{YHOCCY19} were used to augment training image data for wheat head detection, where sampling mask locations were sampled uniformly without any content-aware guidance.
A more relevant study to ours is \mbox{BBoxCut}~\citep{SOB25}, which restricts masks to regions within bounding boxes to occlude parts of wheat heads during detection model training. 
% \color{blue}
% As a result, \mbox{BBoxCut}'s masking is restricted only within the the bounding box and does not allow for masking contextual visual cues. 
% \color{black}
In contrast, our \mbox{CA-Cut} method is designed for visual navigation in mobile robots and probabilistically places masks both \emph{around} and directly on visual cues (i.e.,~crop rows) by sampling from a biased random distribution. 
This setup enables random occlusion of \emph{both} precise keypoints and surrounding ``contextual'' pixels, thereby enhancing robustness to diverse visually challenging scenarios. 
% \color{blue}, without any requirement for a specialized 
% \color{blue}
Similarly, AttMask~\citep{KAGM22} suggests intentionally masking regions where ``teacher'' Vision Transformers have attended while training ``student'' models. 
By comparison, Our method is more general, imposing no requirements on neural network architecture and offering greater computational efficiency, as it does not necessitate training multiple models such as teacher and student. 

\section{CA-Cut}
\label{sec:ca_cut}

% \subsection{Crop-Aligned Cutout Method}
% \subsection{Image Augmentation}
\subsection{Random Masking with Spatial Guidance}
\label{sec:image_augmentation}

Both Cutout~\citep{DT17} and \mbox{CA-Cut} augment images by placing zero-valued masks at random locations. 
Yet, the key distinction is that while the naive version of Cutout~\citep{DT17} is to sample the mask's location uniformly from $\mathcal{U}(0, image\text{ }size)$, \mbox{CA-Cut} employs an attraction force that spatially biases the sampling towards the crop rows. 
As shown in~\autoref{fig:concept}, this design in \mbox{CA-Cut} allows for zeroing out information that is more relevant (e.g.,~crop parts) to visual navigation rather than random content (e.g.,~ground, sky, etc.), thereby promoting the learning of more robust models.

% Our crop-aligned cutout method or CACo 
CA-Cut is simple and intuitive to implement, leveraging labeled images of crop rows for data augmentation. 
For example, the CropFollow dataset~\citep{SMGEVCG21} provides annotated images of cornfields with three keypoints---i.e.,~the vanishing point, and the intercepts of the left and the right crop rows with the image borders---as visualized in~\autoref{fig:concept}. 
In this work, we implement CA-Cut using the lines between the vanishing points and these intercepts. 
For other datasets, alternative labels, such as pixel-wise semantic segmentation~\citep{ACMC21}, may be considered depending on the targeted task. 

For each training image, we randomly select either the left or right side to place a zero-valued mask around the corresponding crop row. 
% , i.e., whether to perform the cutout on the left- or right-hand side of the crop row, with equal probabilities.
Then, a random point~($x_{rnd}, y_{rnd}$) lying on the corresponding line~$\ell$ is sampled and perturbed with an additive white Gaussian noise to center a ($w\times h$) zero-valued mask around it: 
\begin{equation}
    \begin{gathered} 
    x_{mask} = x_{rnd} + \lceil{z_x}\rceil \text{ and} \\
    y_{mask} = y_{rnd} + \lceil{z_y}\rceil \\
    \text{where }
    \begin{bmatrix}
    z_{x} \\
    z_{y} 
    \end{bmatrix}
    \sim \mathcal{N} \left(
    \begin{bmatrix}
    0 \\
    0 
    \end{bmatrix},
    \begin{bmatrix}
    \sigma^2 & 0 \\
    0 & \sigma^2
    \end{bmatrix} \right). 
    \end{gathered}
\end{equation}
$\sigma$~is a hyperparameter to control the proximity of generated masks to the lines, measured in pixels. 
Higher values of~$\sigma$ would lead to masks that are more loosely aligned to the crop rows, whereas lower values keep them more tightly aligned.

For implementation, instead of directly sampling ($x_{rnd}, y_{rnd}$) on line~$\ell$, we introduce a random variable~$\alpha \sim \mathcal{U}(0, 1)$ and compute: 
\begin{equation}
    \begin{gathered} 
    x_{mask} = x_v + \lceil{\alpha(x_{inter} - x_v)}\rceil + \lceil{z_x}\rceil \text{ and} \\
    y_{mask} = y_v + \lceil{\alpha(y_{inter} - y_v)}\rceil + \lceil{z_y}\rceil, 
    \end{gathered}
\end{equation}
where $(x_v, y_v)$ represents the vanishing point, and ($x_{inter}, y_{inter}$) denotes the labeled intercept of the selected line~$\ell$.
Note that $x_{mask}$ and $y_{mask}$ are capped to ensure masks to be located within the image. 

\subsection{Multi-mask Augmentation}
\label{sec:multi-mask}

To learn robust features against severe visual obstructions, we extend augmentation to embed $n$~masks per image, potentially hiding more of the original content to encourage the learning of more robust features. 
However, placing multiple masks exclusively near crop rows may lead to dense clustering in the same regions, reducing the diversity of augmented images. 
To mitigate this, we design \mbox{CA-Cut} to apply $k$~crop-aligned masks and $n-k$~uniformly placed masks, where $1 \leq k\leq n$. 
Similar to Cutout~\citep{DT17}, the uniform placement strategy sets the center of each $w\times h$~mask to: 
\begin{equation}
    (x_{mask}, y_{mask}) \sim \mathcal{U}(\lceil{\frac{w}{2}}\rceil, W-\lceil{\frac{w}{2}}\rceil) \times \mathcal{U}(\lceil{\frac{h}{2}}\rceil, H-\lceil{\frac{h}{2}}\rceil),
\end{equation}
where $W$ and $H$ denote the image width and height, respectively. 
% \mbox{CA-Cut} can be applied repeatedly to an image to generate multiple masks, hiding more of the original content.
In~\autoref{sec:comparative_eval}, we explore the impact of varying~$k$ on overall performance.

\begin{figure}
    \centering
    % \subfloat[]{\label{fig:demo_label}%
    \includegraphics[width=.5\linewidth]{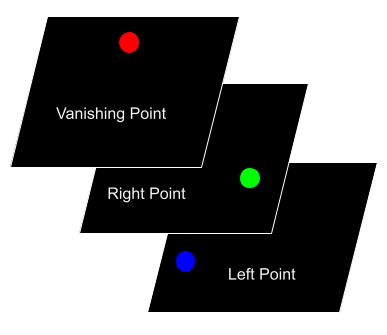}
    % }\\
    % \subfloat[]{\label{fig:true_label}%
    % \includegraphics[width=.8\linewidth]{images/ex.jpg}}
    % \caption{Example of labels used for reconstruction, 
    \caption{
    % \protect\subref{fig:demo_label} 
    Ground truth label visualization with three keypoints, each represented in a separate channel. Each channel contains a value of~$1$ at the keypoint location and $0$ elsewhere.} 
    % demonstration of how each points is represented channel-wise}
    % \protect\subref{fig:true_label} a true label used during training with points exaggerated for example.}
    \label{fig:label_demonstrations}
\end{figure}

\section{Experiments}
\label{sec:experiments}

In this section, we first describe specific experimental configurations, including the dataset, model architectures, hyperparameters, and baselines used, to ensure the results are reproducible. 
Specific results from comparative evaluations will then be presented, accompanied by quantitative and qualitative analyses. 
Lastly, ablation studies will provide justification for our design choices. 

\subsection{Experimental Setup}
\label{sec:experimental_setup}

\subsubsection{Keypoint Prediction \& Dataset}

We used the CropFollow~\cite{SMGEVCG21} dataset to train deep-learing models for the three-keypoint prediction task, introduced in~\citep{sivakumar2024demonstrating}, throughout following experiments. 
The dataset contains a total of $25,296$~color video frame images captured by mobile robot platforms with RGB~cameras while deployed under cornfield canopies in Illinois and Indiana, and each image has a resolution of~$1280\times720$. 
However, we have found that only $1,030$~images are provided with keypoint labels, so we used only these labeled images for our experiments.

Note that while the authors in~\citep{sivakumar2024demonstrating} evaluated the model performance on the estimated robot's heading and the distance ratio between the left and right crop rows based on predicted keypoints, our evaluation focuses directly on the keypoint predictions to assess the performance of trained perception model. 
Specifically, we compute the ``Euclidean distance''  between each predicted keypoint and its corresponding ground truth location in the original image resolution.

% The labels are provided in a csv file that provides the heading angle, distance ratio, and coordinates of 3 points representing the vanishing, left, and right points.

In particular, we also identified \emph{five} distinct sequences within the dataset, presenting various crop growth levels. 
In addition, some consecutive image frames appear highly similar. 
Thus, for each sequence, we used the first $80\%$~images for training and the remaining~$20\%$ for validation so as to minimize the risk of overestimating the performance and ensure evaluations on ``unseen'' video frames.  
% and $20\%$ split was taken for training and validation from each of the sequences. 
% We believed that this method will help the model learn robust features to make predictions, but also provide us with model stability during our experiment. We trained the model with a learning rate of 1e-3 and 32 batch size for 25 epochs, and used the validation set to pick the model with the lowest loss during training.

\begin{figure}[t]\centering
    \subfloat[]{\label{fig:cacut_left_good}%
    \includegraphics[width=.47\linewidth]{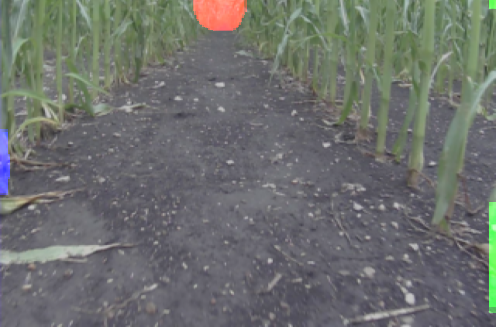}}
    \quad
    \subfloat[]{\label{fig:cacut_right_good}%
    \includegraphics[width=.47\linewidth]{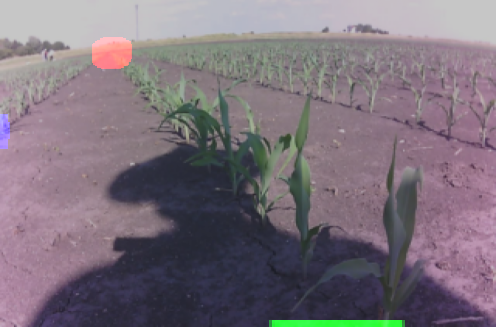}}\\
    \subfloat[]{\label{fig:cacut_van_bad}%
    \includegraphics[width=.47\linewidth]{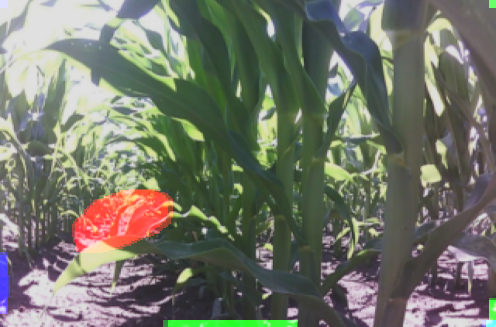}}
    \quad
    \subfloat[]{\label{fig:cacut_right_bad}%
    \includegraphics[width=.47\linewidth]{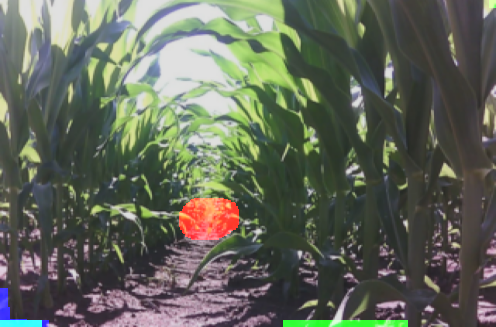}}\\
    \subfloat[]{\label{fig:cutout_outlier}%
    \includegraphics[width=.47\linewidth]{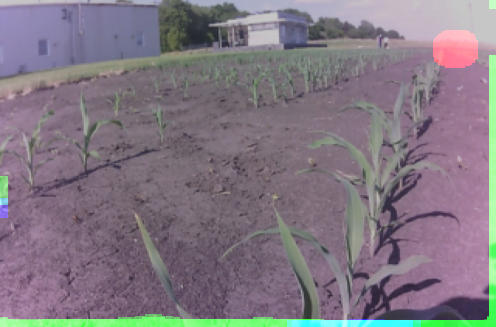}}
    \quad
    \subfloat[]{\label{fig:cacut_outlier}%
    \includegraphics[width=.47\linewidth]{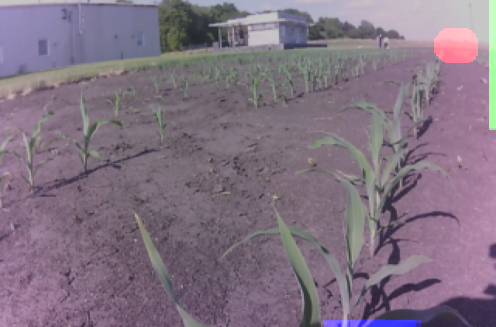}}
        \caption{
            Examples of keypoint prediction heatmaps---red (vanishing point), green (right intercept), and blue (left intercept)---overlaid on input images for visualization. 
            For visibility, each heatmap was enhanced using OpenCV's dilation technique~\citep{OpenCV}, applied twice with a $(5 \times 5)$ kernel. 
            \protect\subref{fig:cacut_left_good}, \protect\subref{fig:cacut_right_good}: \mbox{CA-Cut}'s accurate predictions; \protect\subref{fig:cacut_van_bad}, \protect\subref{fig:cacut_right_bad}: \mbox{CA-Cut}'s inaccurate predictions for right keypoints; 
            % \protect\subref{fig:ca_cut_unseen} presents predictions on a frame sample from a unseen sequence. 
            \protect\subref{fig:cutout_outlier}: An outlier case where Cutout yielded large errors ($>600$) for both left and right keypoints, while \mbox{CA-Cut} in~\protect\subref{fig:cacut_outlier} performed reliably. 
        %     from \mbox{CA-Cut}, with likelihood heatmaps 
        %     \protect\subref{fig:good1} and \protect\subref{fig:good2} represent accurate predictions in complex environments, while \protect\subref{fig:bad1} and \protect\subref{fig:bad2} display relatively poor predictions particularly on vanishing points.  
            }
        \label{fig:prediction_examples}
\end{figure}

\subsubsection{Model Architecture}
\label{sec:model_arch}

To focus only on evaluating the proposed data augmentation technique, we employ an already established model for visual under-canopy navigation with keypoint prediction. 
To be specific, we used a U-Net-based convolutional encoder-decoder architecture that was adopted in \mbox{CropFollow++}~\citep{sivakumar2024demonstrating}.
% The visual navigation model with which we performed our experiment is an encoder-decoder U-Net \cite{ronneberger2015u} architecture proposed by the authors of Cropfollow \cite{sivakumar2024demonstrating}. 
The encoder is constructed with a \mbox{ResNet-$18$} backbone pretrained on ImageNet~\cite{he2016deep} to learn to extract informative visual features into a compressed representation. 
The decoder then utilizes this representation to reconstruct the three keypoints accurately. 
% , with the goal of label reconstruction.

Since the code for \mbox{CropFollow++}~\citep{sivakumar2024demonstrating} is not publicly available, we implemented a similar architecture based on their description. 
Specifically, the encoder is fed a single input image, resized from~$1280 \times 720$ to~$320 \times 240$ and normalized to a $[0, 1]$ range, and consists of four $2$D~convolutional layers with ReLU activation functions, each followed by a max-pooling layer, with an increasing number of $3\times3$ kernels: $64, 128, 256,$ and $512$. 
The decoder has this structure in reverse order, replacing max-pooling layers with upsampling layers to eventually produce a three-channel output image of the same size as the input. 
Additionally, the $3$D feature maps from each encoder are passed to a corresponding layer in the decoder, as designed in conventional U-Nets~\citep{FDLSDJ20}. 

The three output channels correspond to the vanishing point, the right intercept, and the left intercepts, respectively. 
By applying the softmax activation function to each output channel, the model generates spatial likelihood maps for the predicted keypoints. 
For training, ground truth data is represented as a $3$D zero tensor with ones placed only at the corresponding keypoint locations, as illustrated in~\autoref{fig:label_demonstrations}. 

% To fit our reconstruction task we: 1) resized the original image to have the dimensions $320 \times 224$, 2) converted the labels to fit the newly resized images, and 3) generated image labels from the converted labels. The generated image labels have the same dimensions as the resized image. 
% Each color channel only has one value set to 255 to represent one of the three points we want to predict as demonstrated (Fig. \ref{fig:label_demonstrations}). 

\subsubsection{Tested Models \& Training Configurations}

To assess the effectiveness of \mbox{CA-Cut}, the following models are trained and validated. 
Each model is constructed with the identical backbone explained in~\autoref{sec:model_arch} and applies traditional image augmentation techniques (such as color jittering, Gaussian blur, and horizontal flip) by default: 
\begin{itemize}
    \item \textbf{CropFollow++ (CF++):} Use neither Cutout nor CA-Cut but only traditional image augmentations as in~\citep{sivakumar2024demonstrating}. 
    \item \textbf{Cutout: } Apply $n=10$~random masks with no particular spatial distribution guidance~\citep{DT17}.
    \item \textbf{CA-Cut: } Apply $n=10$~random masks, $k$ of which are placed near crop rows using a fixed~$\sigma=100$, while the remaining ones are uniformly sampled.
    % \item \textbf{Mixed: } Apply three random masks by \textbf{Cutout} and two random masks by \mbox{\textbf{CA-Cut w/o Curr.}}. 
    % \item \textbf{CA-Cut: } Apply random masks near crop rows under a ``curriculum'' using $\sigma = \min(25, 100-6t)$ at epoch~$t$ to gradually decrease~$\sigma$ from~$100$ to~$25$. 
\end{itemize}
%
% \subsubsection{Training Procedure}
% To fit our reconstruction task we: 1) resized the original image to have the dimensions $320 \times 224$, 2) converted the labels to fit the newly resized images, and 3) generated image labels from the converted labels. The generated image labels have the same dimensions as the resized image. Each color channel only has one value set to 255 to represent one of the three points we want to predict as demonstrated (Fig. \ref{fig:label_demonstrations}). 
% We implemented a channel-wise cross-entropy loss, where each channel is flattened into a one-dimensional array for both the labels and predictions. We also identified 5 sequences within the dataset, each of which are classified by crop growth levels. These 5 sequences were isolated and an 80-20\% split was taken for training and validation from each of the sequences. We believed that this method will help the model learn robust features to make predictions, but also provide us with model stability during our experiment. 
% \color{blue}
Masks were sampled after each training image was downsized to~$320\times240$. 
% the initial image preprocessing steps—namely, resizing and normalization—which prepare the inputs for the encoder.
% \color{black}
In particular, each mask is of size~$60\times60$.
Note that the specific design choices (e.g.,~mask size, $\sigma$, etc.) are justified in~\autoref{sec:ablation_study}.

In addition, we trained these models with a learning rate of $0.001$ and a batch size of~$32$ for $45$~epochs, using the cross-entropy loss function applied to each channel. 
After each epoch, the model was evaluated on the validation set, and the best-performing model with the minium loss was saved. 
We report the average performance over \emph{five} independent training and validation sessions. 

% the validation set to pick the model with the lowest loss during training.

\begin{table}[]
    \small
    \centering
    % \begin{tabular}{|C{10mm}|C{10mm}|C{10mm}|C{10mm}|C{10mm}|C{10mm}|}
    \begin{tabular}{|C{14mm}||C{9mm}|C{10mm}|C{8mm}|C{8mm}|C{8mm}|C{8mm}|C{8mm}|}
    \hline
    \Large
    \textbf{} & \multirow{3}{*}{\textbf{CF++}} & \multirow{3}{*}{\textbf{Cutout}} & \multicolumn{4}{c|}{\textbf{CA-Cut}} \normalsize \\
    \cline{4-7}
     & & & $k=2$ & $k=5$ & $k=8$ & $k=10$ \\
    \hline
    \hline
    \textbf{Average} & $31.4$ & $25.0$ & $22.9$ & $\mathbf{19.8}$ & $20.3$ & $22.6$ \\
    \hline
    \textbf{Vanishing} & $27.7$ & $14.0$ & $17.4$ & $\mathbf{13.3}$ & $14.0$ & $14.8$ \\ 
    \hline
    \textbf{Left} & $28.8$ & $29.5$ & $24.7$ & $23.0$& $\mathbf{21.4}$ & $24.2$ \\
    \hline
    \textbf{Right} & $37.7$ & $31.7$ & $26.6$ & $\mathbf{23.1}$& $25.5$ & $28.7$ \\
    \hline
    \end{tabular}
    \caption{Semantic keypoint prediction errors.}
    \label{tab:comparison}
\end{table}
    
\subsection{Comparative Evaluations}
\label{sec:comparative_eval}

\Autoref{tab:comparison} shows the average prediction error (i.e.,~the Euclidean distance between ground truth points and predicted ones), using the original image size. 
The significantly poorer performance of \mbox{CF++} \emph{supports} our hypothesis that relying only on traditional image augmentation techniques leads to suboptimal models. 
In constrast, masking-based augmentations, such as Cutout and \mbox{CA-Cut}, improve prediction quality, implying that artificially generating occlusions is well-suited to under-canopy perception. 
In particular, the best-performing \mbox{CA-Cut} model ($k=5$) reduced the error of \mbox{CF++} by $36.9\%$.

Furthermore, compared to using Cutout alone, \mbox{CA-Cut} consistently produced more accurate predictions on average across $k$~values---e.g.,~for example, achieving a~$20.8\%$ average error reduction at $k=5$. 
This result suggests that by ensuring to occlude crop row regions during augmentation, \mbox{CA-Cut} enables the model to learn more \emph{robust} features for keypoint localization. 
Although evaluating the downstream impact on control is beyond the scope of this work, we expect this improvement to translate into even grater gains in ``control'' performance when keypoint detections are used for robot pose estimation, as in~\citep{sivakumar2024demonstrating}, since the $20.8\%$ error reduction applies to each keypoint on average. 

Still, masking \emph{only} around crop rows was not optimal, as \mbox{CA-Cut} with~$k=10$ underperformed compared to the more balanced $k=5$~model. 
As discussed in~\autoref{sec:multi-mask}, this may be because restricting occlusions to crop regions reduces the diversity of the augmented dataset, letting trained models overfit features outside crop row regions.

\begin{figure*}[t]\centering
    \subfloat[]{\label{fig:cutout}%
    \includegraphics[width=.49\linewidth]{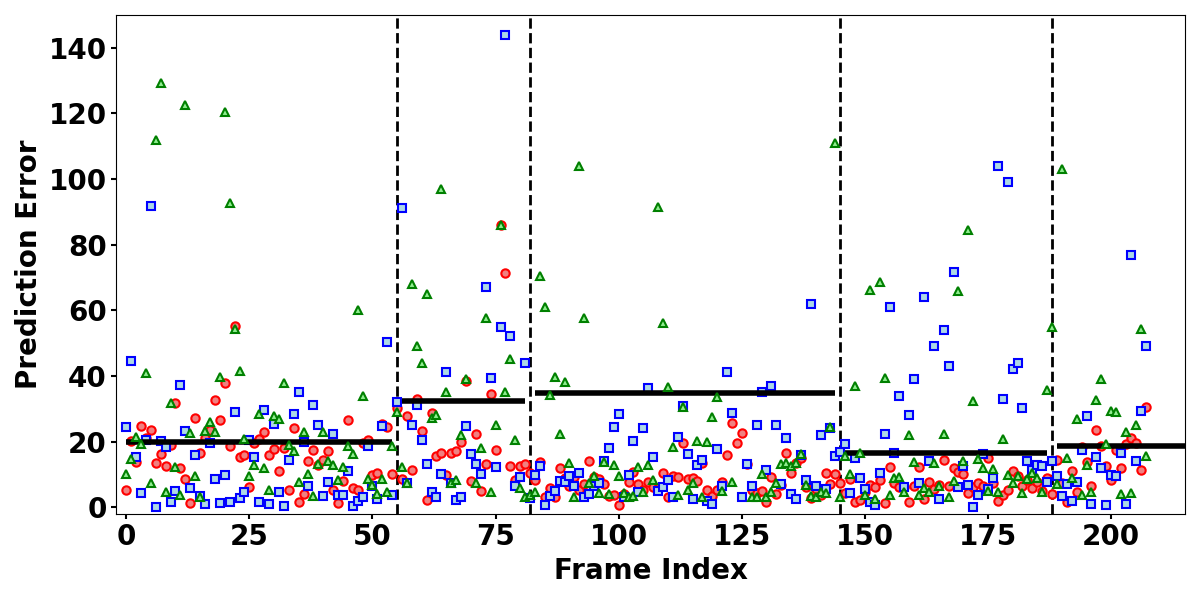}}
    \quad
    \subfloat[]{\label{fig:ca_cut}%
    \includegraphics[width=.49\linewidth]{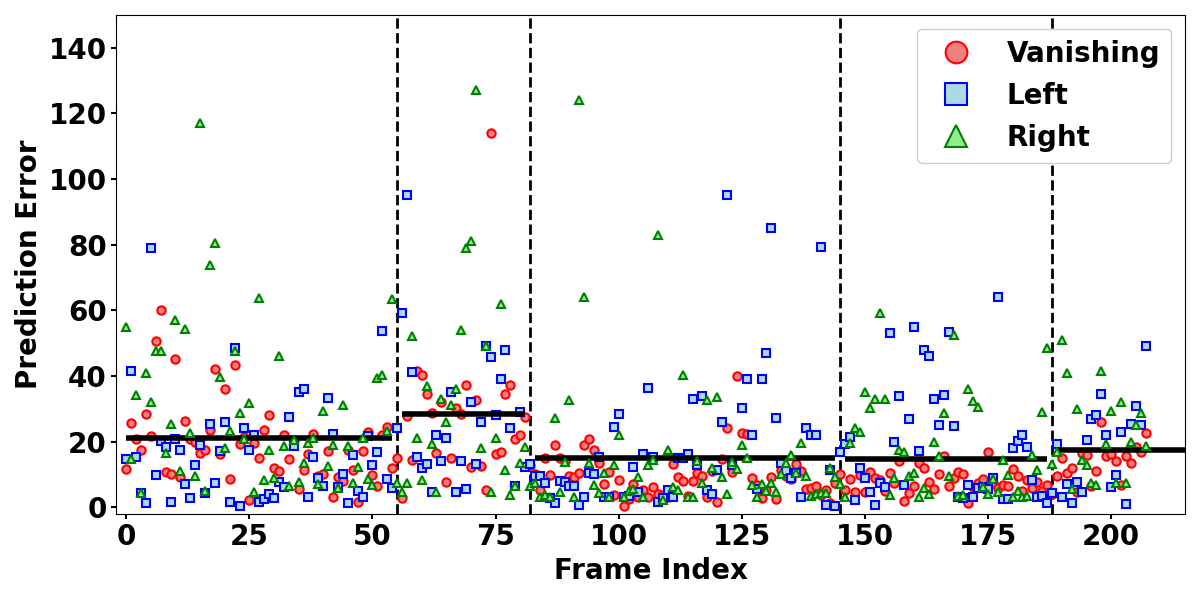}}
        \caption{
            % Validation set prediction error for cutout \protect\subref{fig:cutout} and \protect\subref{fig:ca_cut} with sequence represented by the broken vertical lines and sequence average represented by the broken vertical lines. 
            Video frame-wise prediction errors from \protect\subref{fig:cutout}~Cutout and \protect\subref{fig:ca_cut}~\mbox{CA-Cut}.  Broken vertical lines separate five unique sequences, and each solid horizontal line indicates the average error across all frames within the corresponding sequence.  
            For clarity, six outliers within the range $(350, 1\text{,}100)$---one from sequence~$2$ and five from sequence~$3$---are omitted from~\protect\subref{fig:cutout}. 
            % and sequence average represented by the broken vertical lines.
            % These two plot demonstrate the model train on ca\_cut augmentation \protect\subref{fig:ca_cut} performs well consistently across all sequences while the model trained exclusively on cutout augmentation is less robust and is affect more by changes in environmental variables.
            }
        \label{fig:prediction_error_plots}
\end{figure*}

In addition, \autoref{fig:prediction_error_plots} visualizes frame-wise prediction errors for models trained with Cutout and \mbox{CA-Cut}.
Compared to Cutout, \mbox{CA-Cut} consistently produced lower prediction errors.
Specifically, while Cutout resulted in $22$~predictions with errors exceeding~$80$, \mbox{CA-Cut} had only $11$~such cases.
This observation further supports our claim that perception models trained with \mbox{CA-Cut} augmentations are more \emph{robust}. 
In particular, \mbox{CA-Cut} also exhibited lower error variability across sequences compared to Cutout.  
Cutout produced severe errors ranging from~$350$ to~$1\text{,}100$ for the second and third sequences, whereas \mbox{CA-Cut} maintained prediction errors below $140$ for ``all'' sequences, highlighting its effectiveness in \emph{generalizing} across diverse environments. 

% Although both Cutout and \mbox{CA-Cut} w/o Curr. showed similar performance, the Mixed outperformed all of them, hinting that each technique simulates different types of challenges, ultimately leading to more robust models when combined. 

% Still, \mbox{CA-Cut} demonstrated the \emph{best} performance, highlighting the efficacy of the training curriculum, as observed in~\citep{CWSLC24}.
% In particular, compared to the no-masking baseline (\mbox{CropFollow++}), the error was significantly reduced by $20\%$. 
% This superior result suggests that effective training can be achieved by distributing masks sparsely in early epochs---similar to Cutout---and gradually positioning them more densely around crop rows in later epochs. 
% This strategy may encourage the model to explore diverse visual features across the image during the early learning stages, while preventing ``overfitting'' to local cues from the crops in the later stages. 
% As a result, the model can perform reliably even when nearby plants are occluded or spaced irregularly.  

\subsection{Qualitative Analysis}

% \autoref{fig:prediction_examples} visualizes prediction examples from the model trained with \mbox{CA-Cut} in highly complex environments. 
\Autoref{fig:cacut_left_good} and \autoref{fig:cacut_right_good} show cases where the robot is heading downhill or facing toward the right, which can pose perceptual challenges. 
Nevertheless, the perception models trained with \mbox{CA-Cut} demonstrated reliable performance, accurately detecting all three keypoints.
However, as shown in \autoref{fig:cacut_van_bad} and \autoref{fig:cacut_right_bad}, \mbox{CA-Cut} can still struggle in leafy environments from the second sequence, highlighting the need for further investigation to address such conditions. 

\Autoref{fig:cutout_outlier} illustrates an outlier case for Cutout, where prediction errors exceeded $600$ for both left and right keypoints. 
The model was highly confused due to an extreme viewpoint from the far-left corner, where adjacent rows appeared. 
In contrast, models trained with \mbox{CA-Cut} correctly interpreted the scene and maintained reliable performance (cf.~\autoref{fig:cacut_outlier}).  

% significant occlusion and clutters, hiding frontal views and precise locations of stalks in crop rows, while \autoref{fig:good2} includes farther crop rows, which could potentially confuse the model in intercept predictions. 
% Despite these challenges, the predictions were accurate in both cases, demonstrating the reliability of the predictor. 
% Still, there were other extreme cases, such as in~\autoref{fig:bad1} and \ref{fig:bad2}, where two vanishing points, including the correct one, were identified due to occlusions. 

\subsection{Ablation Study}
\label{sec:ablation_study}

We have first explored various square mask sizes ($w=h$), applying them with Cutout to eliminate any influence of spatially biased mask distribution in \mbox{CA-Cut}.
\autoref{tab:mask_size} shows that both too small~($30$) and too large~($120$) masks degrade performance, and $60\times60$ is the best choice.
The size of~$120$ might erase too much content to learn useful features. 
Hereafter, $60\times60$ will be used for each mask unless mentioned otherwise. 

Next, various numbers of masks~$n$ per image were tested in~\autoref{tab:num_of_masks}, where $10$~masks proved most effective on average. 
Similar to the earlier result, using $20$~masks resulted in the worst performance, obstructing too much critical information in the image.
Although $n=5$ also yielded competitive performance, its vanishing point prediction was significantly worse compared to the model using $10$~masks.  

Lastly, \mbox{CA-Cut} was tested with various~$\sigma$ values (cf.~\autoref{tab:sigma}), when out of $n=10$~masks, $k=5$~masks were crop-aligned (cf.~\autoref{sec:multi-mask}). 
$\sigma=100$~led to the best performance on average. 
In contrast, removing regions too close to the crop rows ($\sigma=50$) resulted in the poorest performance, underscoring the importance of occluding \emph{broader} contextual information to enable robust keypoint predictions even when these surrounding cues are obstructed. 
Moreover, too small~$\sigma$ might lead to densely clustered masks in the same regions, generating similar images and reducing the diversity of augmented images. 

Hence, the design choice used in \autoref{sec:comparative_eval} proved to be optimal. 

% \autoref{tab:mixed} demonstrates that a balance use of Cutout and CA-Cut ($k=2$ or $k=3$) is the most effective strategy. 
% Lastly, the curriculum learning yielded different results depending on the maximum~$\sigma$ value set in the beginning and incrementally decreasing the noise to a minimum~$\sigma$ of 25. 
% Lastly, the outcomes of curriculum learning varied based on the initial maximum~$\sigma$ value, with all converging toward a minimum~$\sigma$ of 25.
% Surprisingly, a curriculum with a range from $100$ to $25$ produced significantly improved performance, especially when compared to the models that statically relied on these $\sigma$ values, as shown in~\autoref{tab:sigma}. 

% From our initial experiment to determine the optimal dimension of a single cutout augmentation square, we discovered that a $60 \times 60$ dimension square (referred to as a 60 dimension cutout square from here on out) improves the performance of the model by 5.77\% over the baseline model, and 4.35\% and 11.5\% over cutout of dimensions of 40 and 70 respectively.
\begin{table}[t]
    \small
    \centering
    \begin{tabular}{|C{15mm}||C{12mm}|C{12mm}|C{12mm}|}
        \hline
        % \multicolumn{5}{|c|}{\textbf{Single-cut}} \\
        % \hline
        % \hline
         & \textbf{30} & \textbf{60} & \textbf{120} \\
        \hline
        \hline
        \textbf{Average} & 26.3 & \textbf{25.0} & 32.0 \\
        \hline
        \textbf{Vanishing} & 22.0 & \textbf{14.0} & 21.9 \\
        \hline
        \textbf{Left} & \textbf{24.0} & 29.4 & 33.0 \\
        \hline
        \textbf{Right} & 32.8 & \textbf{31.7} & 41.1 \\
        \hline        
    \end{tabular}
    % \caption{Average performance, the average $L_{2}$ distance between the ground truth and predicted point, comparison between baseline augmentation and cutout at multiple scales. The baseline has no random erase applied.}
    \caption{Prediction errors for Cutout as the mask size varies, using $10$~masks.}
    \label{tab:mask_size}

    % \begin{tabular}{|c|c|c|}
    
    %     \hline
    %     \multicolumn{3}{|c|}{\textbf{Multi-cut (w/ 5 cuts)}} \\
    %     \hline
    %     \hline
    %     20 x 20 & 40 x 40 & \textbf{60 x 60} \\
    %     \hline
    %     7.09 & 6.30 & \textbf{5.70} \\
    %     \hline
    % \end{tabular}
    % \caption{Average performance comparison of 5-cut cutout augmentation applied at multiple scales.}
    % \label{tab:5_cut}
\end{table}
\begin{table}
    \small
    \centering
    \begin{tabular}{|C{15mm}||C{12mm}|C{12mm}|C{12mm}|}
        % \hline
        % \multicolumn{3}{|c|}{\textbf{60 x 60}} \\
        % \hline
        \hline
          & \textbf{5} & \textbf{10} & \textbf{20} \\
        \hline
        \hline
        \textbf{Average} & 26.0 & \textbf{25.0} & 26.9 \\
        \hline
        \textbf{Vanishing} & 20.8 & \textbf{14.0} & 21.9 \\
        \hline
        \textbf{Left} & \textbf{25.8} & 29.4 & 27.1 \\
        \hline
        \textbf{Right} & \textbf{31.3} & 31.7 & 32.0 \\
        \hline 
    \end{tabular}
    % \caption{Average performance comparison of a $60 \times 60$ scale cutout augmentation applied with 3, 5, and 7 cuts per image.}
    \caption{Prediction errors for Cutout as the number of ($60\times60$) masks~$n$ changes.}
    % with the mask size of~$60$.}
    \label{tab:num_of_masks}

\end{table}
% , \ref{fig:bad1}, and \ref{fig:bad2}
% We chose 3 candidates, the 20, 40 and 60, dimension cutouts and tested the average performance of each model across 5 runs. As determined in the single-cut experiment, we also found the 60 dimension cutout square to be optimal for multi-cut cutout augmentation, showing a performance 9.5\% better than the second-best model, trained with the 40 dimension cutout square.
\begin{table}[!h]
    \small
    \centering
    
    \begin{tabular}{|C{15mm}||C{12mm}|C{12mm}|C{12mm}|}
        % \hline
        % \multicolumn{3}{|c|}{\textbf{Informed Multi-cut (w/ 5 cuts and Gaussian noise $\sigma$)}} \\
        % \hline
        \hline
         & \textbf{50} & \textbf{100} & \textbf{200} \\
        \hline
        \hline
        \textbf{Average} & 24.2 & \textbf{19.8} & 21.2 \\
        \hline
        \textbf{Vanishing} & 15.5 & \textbf{13.3} & 19.7 \\
        \hline
        \textbf{Left } & 32.5 & 23.0 & \textbf{19.5} \\
        \hline
        \textbf{Right } & 24.6 & \textbf{23.1} & 24.4 \\
        \hline
    \end{tabular}
    \caption{Prediction errors for \mbox{CA-Cut} with five masks, randomly sampled using different~$\sigma$ values, out of ten~$(60\times60)$ masks.}
    % with five masks of size~$60$.}
    \label{tab:sigma}
\end{table}

\section{Limitations \& Future Work}
\label{sec:limit}

% one data/one type 
% While our work demonstrates the potential of CA-Cut augmentation for improving visual navigation model training, several limitations should be acknowledged.
While \mbox{CA-Cut} has been designed as a generically applicable data augmentation technique, our experiments were limited to the CropFollow dataset~\citep{SMGEVCG21, sivakumar2024demonstrating}. 
For future work, we plan to validate our technique on other datasets, including those for sugar beet~\citep{DCWG24} and vineyard~\citep{ACMC21}, and to explore its applicability to over-canopy and orchard navigation scenarios, as well as to semantic segmentation tasks.  
% We did not evaluate CA-Cut augmentation on other dataset types, such as those designed for segmentation-based visual navigation models. As a result, the generalizability of our findings to other forms of visual navigation remains an open question for future work, although we believe that it can be generalized from our experience and literature review.

% control 
% First, we are grateful to \citet{sivakumar2024demonstrating} for providing a clean and well-organized dataset, which was instrumental to our experiments. 
Unlike \mbox{CropFollow++}~\citep{sivakumar2024demonstrating}, where detected keypoints were translated into the robot's heading and the distance ratio between left and right crop rows, our experiments focused solely on the vision component. 
This choice was primarily because the robot's ``roll'' information captured by the inertial measurement unit~(IMU) is unavailable in the dataset, even though it was a key input in their framework.
To conduct physical experiments, we will either collect a custom dataset containing all essential information or implement a controller similar to the Segmentation to Proportional Control (SPC) model~\citep{ACMC21}, which does not rely on external IMU data. 

% for this conversion.
% However, our reliance on this specific dataset imposed constraints on the evaluation metrics we could use. In particular, we were limited to using the Euclidean distance between ground truth and predicted points. Although the authors offer a detailed explanation of their keypoint-to-heading estimation pipeline, we were unable to replicate this aspect due to the unavailability of certain data, such as IMU readings and configuration details related to data collection. 

% field evaluation
% 
%%%

% For future work, we plan to explore more appropriate metrics for error estimation than Euclidean distances in pixel space.  
% This is because pixel space distances can be misleading, as they depend on the image size. 
% In addition, we will evaluate the robot's heading and distance ratio estimation, following the approach in~\citep{sivakumar2024demonstrating}, to assess the reliability of perception for robot motion control. 

% In particular, we will perform field evaluations, deploying the trained perception models on physical robots to examine the reality gap. We also aim to explore alternative methods for converting keypoints into heading and distance ratios for robot motion control. 
% Furthermore, we plan to investigate the applicability of \mbox{CA-Cut} to other crop types and different tasks.
% For instance, beyond under-canopy navigation, we could consider over-canopy navigation scenarios. 
% Similarly, we aim to evaluate its performance in other relevant tasks, such as semantic segmentation, instead of keypoint predictions. 

\section{Conclusion}

We have introduced \mbox{CA-Cut}, a spatially guided augmentation technique designed to train robust perception models for visual under-canopy navigation in complex agricultural environments with variable row spacing and occlusions. 
Through keypoint prediction tasks, we demonstrated the effectiveness of masking-based image augmentations in simulating visual obstructions and improving model robustness. 
Our results show that biasing the mask distribution toward crop rows in \mbox{CA-Cut} is critical for enhancing both prediction accuracy and generalizability across diverse environments---achieving up to a $36.9\%$~reduction in prediction error.  
We also discussed the limitations of this work and future research directions to further enhance model reliability and robustness. 

\section*{Acknowledgment}
This work was partially supported by NSF~(2300955) and by USDA-NIFA~(2023-38640-39572) through the Southern SARE program~(GS24-301).

\addtolength{\textheight}{-12cm}   % This command serves to balance the column lengths

{\small
    \ifx\usenatbib\undefined%
	\bibliographystyle{IEEEtran}%
    \else%
    \bibliographystyle{IEEEtranN}%
    \fi
	%\bibliography{IEEEabrv, IEEEexample}
	\bibliography{references}
}

\end{document}